\documentclass[conference]{IEEEtran}
\IEEEoverridecommandlockouts
\usepackage{cite}
\usepackage{amsmath,amssymb,amsfonts}
\usepackage{algorithmic}
\usepackage{graphicx}
\usepackage{textcomp}
\usepackage{float}
\usepackage{xcolor}
\def\BibTeX{{\rm B\kern-.05em{\sc i\kern-.025em b}\kern-.08em
    T\kern-.1667em\lower.7ex\hbox{E}\kern-.125emX}}
    
\begin{document}

\title{Machine Learning-Assisted Analysis of Small Angle X-ray Scattering}

\author{\IEEEauthorblockN{Piotr Tomaszewski,
Shun Yu, Markus Borg and
Jerk Rönnols}
\IEEEauthorblockA{RISE Research Institutes of Sweden, Lund and Stockholm, Sweden\\
Email: \{piotr.tomaszewski, shun.yu, markus.borg, jerk.ronnols\}@ri.se}}
\maketitle

\begin{abstract}
Small angle X-ray scattering (SAXS) is extensively used in materials science as a way of examining nanostructures. The analysis of experimental SAXS data involves mapping a rather simple data format to a vast amount of structural models. Despite various scientific computing tools to assist the model selection, the activity heavily relies on the SAXS analysts' experience, which is recognized as an efficiency bottleneck by the community. To cope with this decision-making problem, we develop and evaluate the open-source, Machine Learning-based tool SCAN (SCattering Ai aNalysis) to provide recommendations on model selection. SCAN exploits multiple machine learning algorithms and uses models and a simulation tool implemented in the SasView package for generating a well defined set of datasets. Our evaluation shows that SCAN delivers an overall accuracy of 95\%-97\%. The XGBoost Classifier has been identified as the most accurate method with a good balance between accuracy and training time. With eleven predefined structural models for common nanostructures and an easy draw-drop function to expand the number and types training models, SCAN can accelerate the SAXS data analysis workflow. 
\end{abstract}


\section{Introduction}
The advancing of our society is greatly related to the development of new materials. From the stone age to the silicon era, even till the coming quantum world, new materials lay the groundwork for novel technologies. All materials are built upon controlled arrangement of atoms via either naturally occurring or artificially imposed design. Such an arrangement is known as materials structure, which usually demonstrates a hierarchical characteristic. From an \AA ngstrom, about the size of one atom, to kilometers which represents the largest sizes of man-made objects, materials may be shaped into different structures at different levels to serve its intended functions. 

Nanotechnology has revolutionized materials science. Since the famous quote \textit{"There’s Plenty of Room at the Bottom"} by Dr. Richard Feynman \cite{feynman2018there}, efforts to control materials structure have been focused on the nanoscale -- this is now widely known as nanomaterials. Materials with designed nanostructures can demonstrate properties that would never be possible through conventional material design.

Advanced characterization methods play a central role in the development of new materials. At research infrastructures such as the Swedish national MAX IV Laboratory, a 4th  generation synchrotron radiation facility, X-rays are used to get unique insights into materials in terms of structure and properties\cite{MAXIVlink}. The obtained knowledge paves the way for the development of new materials in various fields, e.g., medications, microelectronics, energy generation and storage, biodegradable plastics, and novel packaging, etc.\cite{fan2018synchrotron} In general, the synchrotron radiation based techniques could be classified into three categories: i) scattering, ii) spectroscopy, and iii) imaging, which provide complimentary information of the material structure and properties\cite{willmott2019introduction}.

Scattering is powerful in structure determination from submicron down to \AA ngstrom. It is a process where incoming X-ray photons are forced to change the travelling direction to a certain angle, known as scattering angle, after an interaction with materials. The intensity of the scattered X-ray as a function of scattering angle contains information of the materials structure. Theoretically, the intensity distribution follows the magnitude of Fourier transformation (FT) of the electron density distribution in the materials. This leads to a fact that the size which scattering technique could probe is inversely proportional to the scattering angle. Thus, practically, scattering characterization is performed at two geometries depending on the scattering angle, i.e. small angle X-ray scattering (SAXS) and wide angle X-ray scattering (WAXS). While WAXS is more sensitive to the local atomic structure, SAXS experiments cover the material structures spanning over several orders of magnitude from several hundred nanometer to several nanometer by a single scattering pattern. In applications, SAXS has been used to study various nanostructures' shapes, which critically relates to material properties. Furthermore, the high photon flux at a synchrotron radiation facility allows SAXS experiments to be performed at tens Hz up to thousand Hz with a single frame of $\sim$MB size. Abundant SAXS data can even serve as a real-time animation revealing a continous nanostructure development. 

However, mapping of SAXS data to real nanostructures is not straightforward. SAXS intensity is essentially the magnitude of FT without ``phase'' information. Experimentalists use SAXS tools to match each single experimental data set with a set of existing models representing the studied nanostructure. As more high data-rate SAXS experiments are foreseen, such traditional analysis approaches could become a bottleneck in the process. The tools offer a large selection, sometimes hundreds, of predefined models to choose from. There is no easy way to know which one to choose, and checking all of them is infeasible. So, effectively, there is a considerable amount of experience necessary to efficiently select the correct model or to even narrow the selection to some plausible model candidates. As pointed out in \cite{archibald2020classifying, do}, this step can be very intimidating to both new and advanced users. 

There are some reports (e.g. \cite{archibald2020classifying, do}) that show encouraging results from using supervised learning to facilitate the model selection. As a part of this study, we have produced an open-source scattering analysis tool called SCAN. We used the tool to evaluate a selection of machine learning (ML) models on our scattering data. The results we have obtained are promising. On our data set, our classification models outperform previous work reported in the literature and the accuracy figures we obtain make them highly relevant to users analyzing SAXS data. We envision that supervised learning could be a practical tool to increase the level of automation in the work of SAXS analysts~\cite{parasuraman}, speeding up the characterization.

The reminder of the paper is structured as follows. Section~\ref{sec:rw} summarizes related work. Section~\ref{sec:tool} presents our open source SCAN tool. The evaluation method is presented in Section~\ref{sec:method} and evaluation results are presented in Sections~\ref{sec:res}. The findings are discussed and compared to related work in Section~\ref{sec:disc} and Section~\ref{sec:conc} presents the conclusions from our work.

\section{Related work} \label{sec:rw}
ML enters new applications areas almost daily, ranging from character recognition, through product recommendations to self driving cars. A surge of applications of ML in scientific computing is also evident~\cite{doi:10.1021/acs.infocus.7e4001, RevModPhys.91.045002, jumper2021highly}. In the field of materials characterization, where large volume multidimensional data are available, the application of ML has a great potential and is clearly gaining momentum\cite{schmidt2019recent}. Recently, Chen et al~\cite{chen} has reviewed the latest progress in the area of X-ray scattering characterization. As an important technique, SAXS has already become a test field for ML-assisted analytics. One type of application is around the experimental configuration; Wang, et al., use deep learning to classify the experimental configuration-based scattering image \cite{wang}. Herck et al, use deep learning to study the structure distribution at grazing incidence geometry~\cite{herck}. Another type of application is about SAXS for biomarcomolecules, which serves a well defined user community. For example, Franke et al., focused on feature extraction of the scattering pattern and use weighted k Nearest Neighbour (wKNN) and wkNN with Gaussian processes to classify low resolution molecular structure into different shapes~\cite{franke2018machine}. He et al., instead turned to the auto-encoder to reconstruct biomacromolecules from bio-SAXS curves~\cite{he2020model}. For the general nanostructure analysis, Archibald et al. used KNN to classify general structural models and analyzing SAXS curves\cite{archibald2020classifying}. Do et al. instead discovered that the Random Forest is a feasible model with an accuracy of 0.783 for such classifications~\cite{do}. 
In this work, along the line of pioneer work of Archibald et al. and and Do et al., we aim to offer materials scientists a tool which \textit{\textbf{synergizes}} different ML algorithms to help SAXS analysts quickly identify a likely model structure for further analysis. 

\section{SCAN tool} \label{sec:tool}
To facilitate SAXS analysts' model selection we have implemented a tool called SCAN (SCattering Ai aNalysis). The tool offers ML model management and classification interfaces. The former makes it possible to train, evaluate and maintain any number of ML classification models. The latter provides possibility to use the tool for performing classification on an experimental data using any of the available classification models. The source code and the training data (described in Section~\ref{sec:data}) are hosted on GitHub~\cite{scan} under an MIT license.

SCAN supports nine different ML classifiers as well as a possibility to run them in a stacked manner. The classifiers included are the following:
\begin{itemize}
    \item Random Forest Classifier
    \item XGBoost Classifier
    \item XGBoost Random Forest Classifier
    \item Multi-layer Perceptron classifier
    \item Decision Tree Classifier
    \item AdaBoost Classifier
    \item Gaussian Naive Bayes
    \item Quadratic Discriminant Analysis
    \item kNeighbors Classifier 
\end{itemize}
All but the two XGBoost classifiers are implemented using the scikit-learn library \cite{scilearn}. XGBoostClassifier and XGBoostRandomForestClassifier come from the XGBoost library \cite{xgboost}. One notable difference between the two libraries is that, at the time of writing, only XGBoost offered GPU acceleration. All algorithms have been run with default parameters, one of the goals with this evaluation was to check if further hyperparameter tuning would be required. 

Except from training and maintaining the classification models, the management interface offers a possibility to reduce the dimensionality of the input data by using Principal Component Analysis (PCA) and to request a k-fold cross validation.

In the classification interface the data is provided in the form of a comma-separated values (CSV) file. The output is another CSV file, where each input item is annotated with a predicted nanostructure shape and a corresponding probability (confidence) for each selected ML classifier. The interaction with the software is either through a command line or through a web interface.

We developed SCAN to meet expectations of a sustainable ML-based tool in scientific computation. During development, we aligned our work with contemporary discussions on the trending topic of MLOps, i.e., standardization and streamlining of ML life cycle management. In line with work on sustainable MLOps in scientific computing by Tamburri~\cite{tamburri2020sustainable}, we supported experiment tracking and explainability by connecting the service Neptune AI~\cite{neptune} to a continuous integration pipeline orchestrated by GitHub Actions. In practice, each pull request (e.g., modifying code, data, or hyperparameters) triggered a reevaluation of model performance and presented the results in an online dashboard. 

\section{Method} \label{sec:method}
This section first presents how we generated representative data using the SASView\cite{sasview}. Second, we present how we trained and evaluated classification models.

\subsection{Data Generation} \label{sec:data}
\begin{figure}
    \centering
    \includegraphics[width=8.5cm]{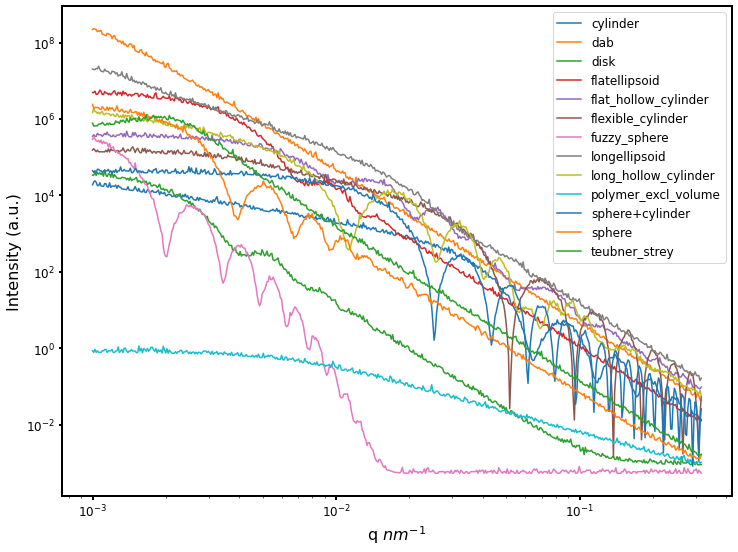}
    \caption{Illustrations of simulated SAXS curves for each model.}
    \label{fig:saxs}
\end{figure}

As aforementioned, ideal scattering intensity is a result of the FT of the electron density distribution inside materials. Thus, SAXS data could be generated via theoretical simulation to meet requirements on large data sets. Here, we use sasmodel module from the SASView software\cite{sasview}, which has been used in previous work\cite{archibald2020classifying, do}. Poisson error distribution has been added onto the generated data to take into account the photon counting event. We have chosen both geometric models and a statistic model to generate the training and test data. For the geometric model, eight common models for nano-objects in solutions were chosen, including sphere, sphere with fuzzy shell, ellipsoid with different aspect ratios, long and/or hollow cylinder, and flat disk. For statistical models, which are common for systems without defined shapes, three models were selected: Debye-Anderson-Brumberger (DAB) Model\cite{debye1957scattering}, Polymer Excluded Volume model\cite{hammouda1993sans} and Teubner-Strey model of microemulsions\cite{teubner1987origin}. Furthermore, we added one mixture case consisting of certain sphere and cylinder with arbitrary weighted non-zero contribution to further challenge the classification process. In each model, the randomized model parameter, such as radius and length, as well as the polydisperse size and aspect ratio were introduced with the methods available via sasmodels and/or Numpy packages. All the data are generated as (x,y) curves with y as the scattering intensity and x as the scattering vector q ($nm^{-1}$), which is defined as $q=\frac{4\pi}{\lambda}sin(\theta)$ with $\theta$ as half of the scattering angle. The q range is chosen $[10^{-3} nm^{-1}, 3 nm^{-1}]$, covering the common SAXS probing range. Selected plots from each model are shown in Fig.~\ref{fig:saxs}. For each model, 3,000 curves were generated and both geometric and statistical model were fed into SCAN for training at the same time. All data are publicly available in the GitHub repository.

\subsection{Modeling and Evaluation}
The candidates for evaluation have been selected in the following way:

\begin{enumerate}
   \item We have trained models for all nine individual classifiers supported by the SCAN tool. On top of that we have trained two stacked models
   \begin{itemize}
     \item StackedAll -- consisting off all individual classifiers
     \item StackedTop5 -- consisting of the five classifiers with the best individual performance
   \end{itemize}
   \item For every candidate classifier, we have considered the following inputs
   \begin{itemize}
     \item all 500 measurement points
     \item PCA set to explain 99\% and 95\% of variance (resulting in 40 and 10 PCA components, respectively)
    \end{itemize}
\end{enumerate}

In total we evaluate 33 candidate classification models. The stacked models are included in the study to investigate if combining multiple classifiers yields a better result that any single classifier. PCA is included for two reasons: it should reduce the training time and there are studies (e.g. \cite{howley}) that report potential positive impact of PCA on the model accuracy. The evaluation follows the same method as used in \cite{do}. We perform 5-fold cross validation and, for each model, we report the mean accuracy together with the standard deviation. k-fold validation is a common method of evaluating classification models, and using the same split parameter as in \cite{do} makes it possible to compare the results. Additionally, to visualize the results, we present confusion matrices for selected models. 

\section{Results} \label{sec:res}
Table~\ref{table:1} summarizes the accuracy numbers accompanied by standard deviation. Individually, XGBoost Classifier and Random Forest Classifier outperform other individual candidates. This holds true regardless if PCA has been applied. The stacked models offer even slightly better results, and the model based on the top 5 individual classifiers is performing at least as well as the one combining all.

\begin{table}
\caption{5-fold prediction accuracy. In brackets stddev.}
\centering
\begin{tabular}{|l|c|c|c|}
 \hline
 \textbf{ML Classifier} & \textbf{All} & \textbf{PCA99} & \textbf{PCA95} \\
 \hline
 XGBClassifier & 0.959 (0.002) & 0.935 (0.003) & 0.901 (0.004) \\
 RandomForestClassifier & 0.958 (0.002) & 0.916 (0.005) & 0.893 (0.005)\\ 
 KNeighborsClassifier & 0.897 (0.002) & 0.900 (0.003) & 0.882 (0.003) \\
 XGBRFClassifier & 0.825 (0.005) & 0.843 (0.004) & 0.750 (0.004) \\
 MLPClassifier & 0.750 (0.012)  & 0.766 (0.009) & 0.740 (0.006)  \\
 QuadraticDA & 0.656 (0.007) &  0.658 (0.007) & 0.625 (0.005) \\
 DecisionTreeClassifier & 0.578 (0.021) & 0.754 (0.005) & 0.680 (0.005) \\
 GaussianNB & 0.471 (0.005) & 0.436 (0.001) & 0.419 (0.003) \\
 AdaBoostClassifier & 0.274 (0.007) & 0.283 (0.021) & 0.282 (0.026) \\
 \hline
 StackedAll & 0.973 (0.002)&  0.948 (0.003)  & 0.925 (0.003) \\
 StackedTop5 & 0.972 (0.002) & 0.948 (0.003) & 0.942 (0.004) \\ 
 \hline
\end{tabular}
\label{table:1}
\end{table}

\begin{figure}

\begin{minipage}{1.0\linewidth}
  \centering
  \centerline{\includegraphics[width=8.5cm]{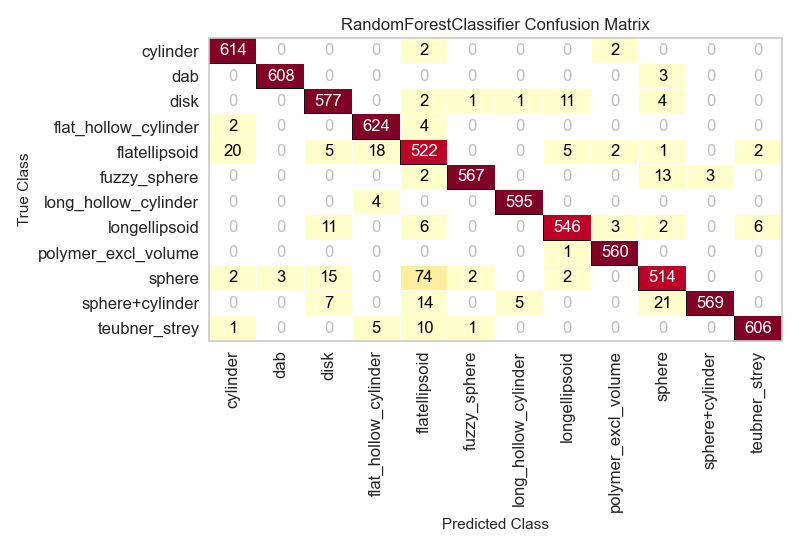}}
%  \vspace{2.0cm}
  \centerline{(a) Random Forest Classifier}\medskip
\end{minipage}

\begin{minipage}[b]{1.0\linewidth}
  \centering
  \centerline{\includegraphics[width=8.5cm]{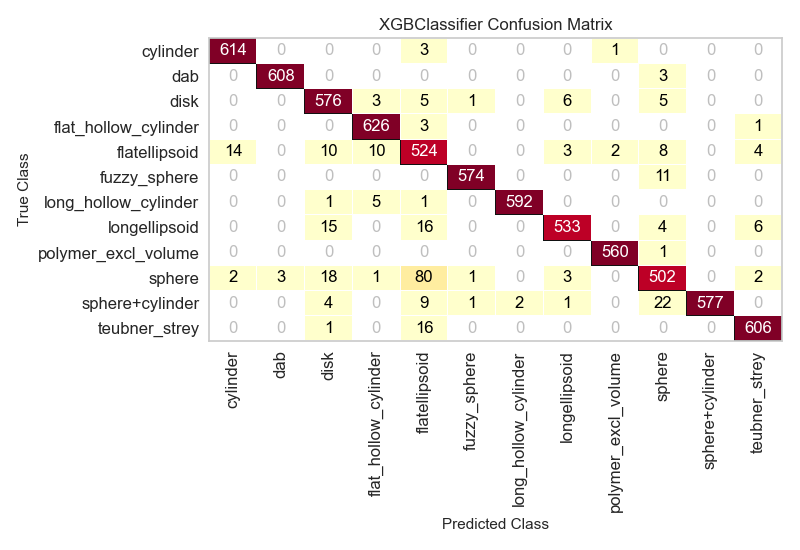}}
%  \vspace{2.0cm}
  \centerline{(b) XGBoost classifier}\medskip
\end{minipage}
\begin{minipage}[b]{1.0\linewidth}
  \centering
  \centerline{\includegraphics[width=8.5cm]{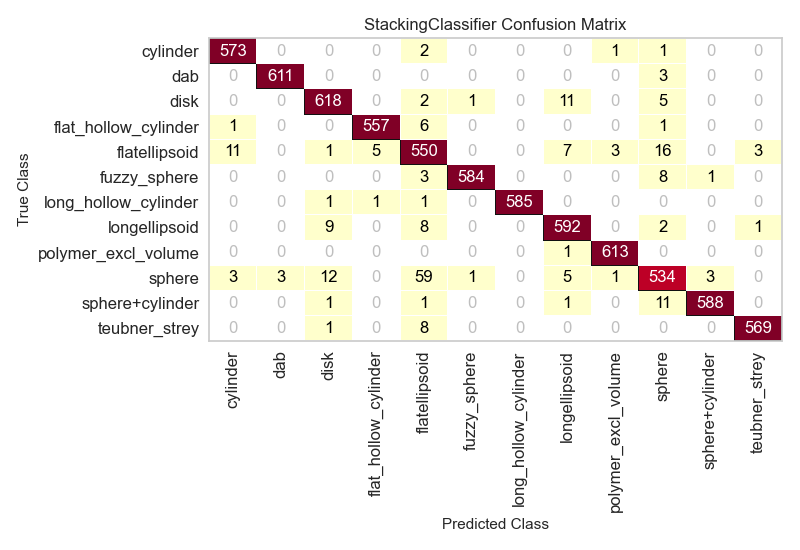}}
%  \vspace{2.0cm}
  \centerline{(c) Stacked - top five classifiers}\medskip
\end{minipage}

\caption{Confusion matrices for top performing models (all inputs).}
\label{fig:res}
\end{figure}

\begin{figure}

\begin{minipage}[b]{1.0\linewidth}
  \centering
  \centerline{\includegraphics[width=8.5cm]{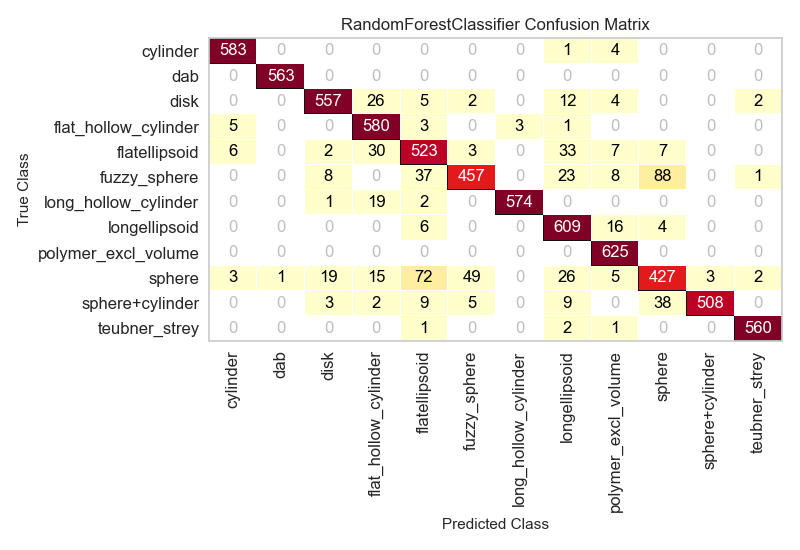}}
%  \vspace{2.0cm}
  \centerline{(a) Random Forest Classifier}\medskip
\end{minipage}

\begin{minipage}[b]{1.0\linewidth}
  \centering
  \centerline{\includegraphics[width=8.5cm]{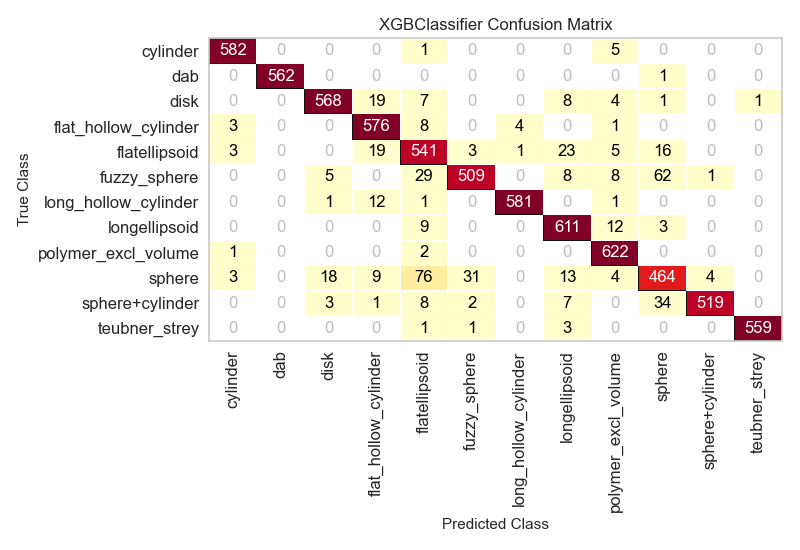}}
%  \vspace{2.0cm}
  \centerline{(b) XGBoost classifier}\medskip
\end{minipage}
\begin{minipage}[b]{1.0\linewidth}
  \centering
  \centerline{\includegraphics[width=8.5cm]{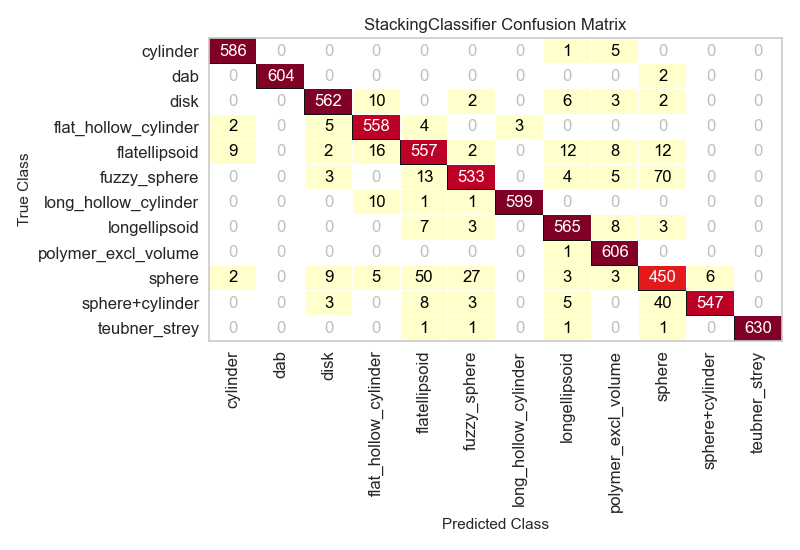}}
%  \vspace{2.0cm}
  \centerline{(c) Stacked - top five classifiers}\medskip
\end{minipage}

\caption{Confusion matrices for top performing models (PCA of 99\%, 40 components).}
\label{fig:res2}
\end{figure}

\begin{figure}
\begin{minipage}[b]{1.0\linewidth}
  \centering
  \centerline{\includegraphics[width=8.5cm]{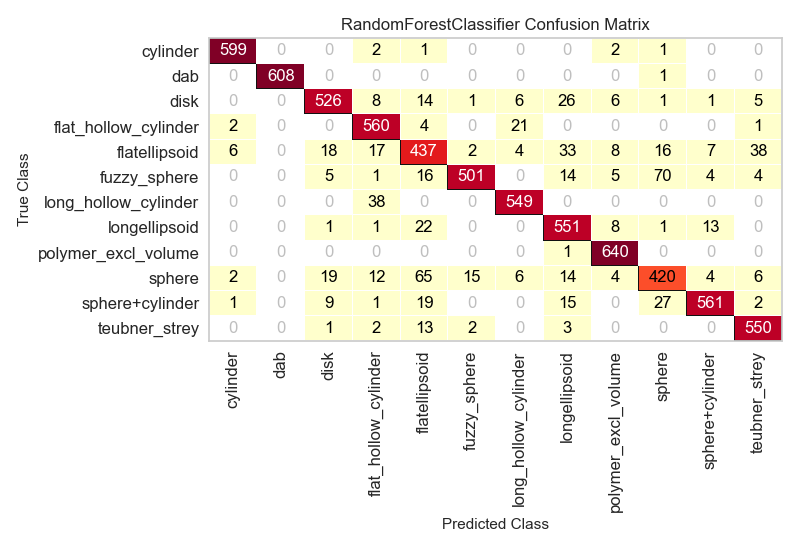}}
%  \vspace{2.0cm}
  \centerline{(a) Random Forest Classifier}\medskip
\end{minipage}

\begin{minipage}[b]{1.0\linewidth}
  \centering
  \centerline{\includegraphics[width=8.5cm]{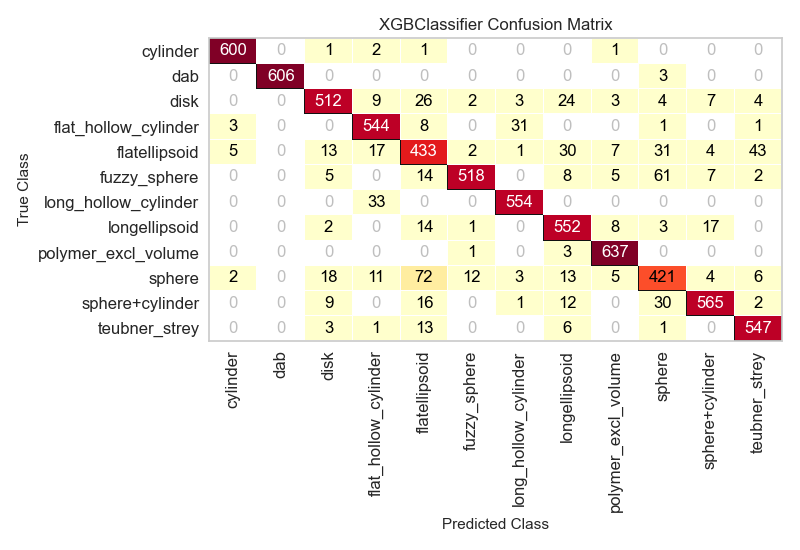}}
%  \vspace{2.0cm}
  \centerline{(b) XGBoost classifier}\medskip
\end{minipage}
\begin{minipage}[b]{1.0\linewidth}
  \centering
  \centerline{\includegraphics[width=8.5cm]{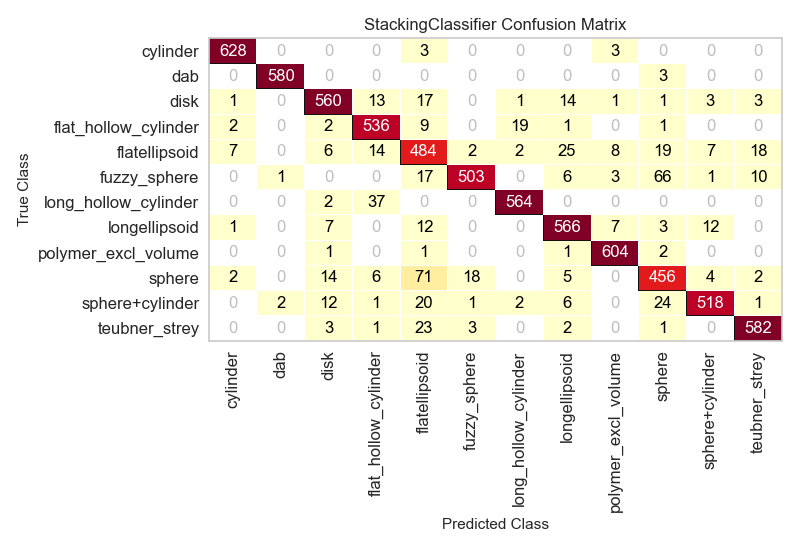}}
%  \vspace{2.0cm}
  \centerline{(c) Stacked - top five classifiers}\medskip
\end{minipage}

\caption{Confusion matrices for top performing models (PCA of 95\%, 10 components).}
\label{fig:res3}
\end{figure}

To closer inspect the performance of the models we have created confusion matrices for the three most promising models for respective inputs, i.e., i) all inputs (Fig. \ref{fig:res}), ii) PCA set to 99\%, and iii) PCA set to 95\% (Fig. \ref{fig:res2} and \ref{fig:res3}, respectively). In the process we have also clocked the training and evaluation for all the models. Table \ref{table:2} presents the timings of the top 3 classifiers depicted in Fig.~\ref{fig:res}-\ref{fig:res3}.

% \usepackage{multirow}
\begin{table}
\caption{Training and evaluation times}
\begin{center}
\begin{tabular}{|l|l|l|l|}
\hline
\textbf{ML Classifier }                         & \textbf{All}     & \textbf{PCA99} & \textbf{PCA95} \\ \hline
XGBClassifier                                    & 0:01:45 & 0:00:46  & 0:00:38  \\ \hline
RandomForestClassifier                           & 0:04:55 & 0:01:15  & 0:00:43  \\ \hline
StackedTop5                                      & 1:05:54 & 0:41:32  & 0:26:11  \\ \hline
\end{tabular}
\end{center}
\label{table:2}
\end{table}

\section{Discussion} \label{sec:disc}

We find our results encouraging. In (Table~\ref{table:1}) we can see that our classification models can reach the accuracy of over $97\%$ for the stacked models. However, the training of these models is the most time consuming - Table~\ref{table:2} shows that their training takes roughly order of magnitude longer time compared to the best individual classifiers, i.e. RandomForrestClassifier and XGBoostClassifier. These individual classifiers do not fall far behind with respect to accuracy, they score almost 96\%. As a result of this observation these two individual classifiers have been selected as default choice in the SCAN tool, leaving the other ones optional.

A close analysis of the confusion matrices reveals that generally lower accuracy predictions are made for sphere, ellipsoidal and sphere-like models -- this phenomenon is consistent across classifiers. We hoped to tackle this by stacking classifiers into an ensemble of diverse models. However, the only marginally better confusion matrices of stacked models confirmed that sphere-like model remains a challenging case. This finding is  consistent with previous work~\cite{archibald2020classifying, do}. Somewhat contradictory, sphere-like shapes is probably one of the most recognizable models in the SAXS textbook\cite{guinier1955small}. One explanation could be that spherical models easily resemble each other upon the common ``sphere''-shape envelope profile and polydispersity of its radius. To better distinguish sphere-like models could certainly be a future topic in ML-assisted scattering analysis. In addition, as an attempt, we also introduced the ``sphere + cylinder'' class by summing cylinder models and sphere models in a certain ratio. As a result, the ``sphere + cylinder'' class obtains better prediction accuracy than the pure sphere-like class. This opens doors to identify the possible shape transition for many soft matter systems during in situ studies\cite{ma13030752}. The mixing model is also an efficient method for data augmentation to expand the model selections. 

We have also experimented with applying PCA. For the best performing classifiers, the PCA application consistently leads to slightly worse accuracy across all tested cases, but at the same time it speeds up the training process. Applying PCA does not change the list of most promising classification models, and their relative performance remains the same. Therefore, should the training time ever become an issue (e.g. extensive experimentation), one may consider using PCA, at least in the experimentation phase, to find the most suitable classification models.

As we considered the obtained accuracy adequate for the practical needs, we have not attempted any hyperparameter tuning in this study. This is an obvious candidate for further work, as it may possibly yield even better results.   

\section{Conclusions} \label{sec:conc}
The purpose of this study was to investigate the applicability of ML to assist SAXS analysts in the task of model selection. In the course of the study, we have developed an open-source tool called SCAN that made it possible for us to create and evaluate eleven ML classifiers trained and evaluated on our scattering data. The tool also provides the classification interface that is meant to provide actual model recommendations to practitioners. 

We have found that on our scattering data multiple ML methods provided satisfactory accuracy for 11 common SAXS models. Random Forest and XGBoost are the most promising classifiers, deeming the accuracy from 89\% to as high as 95.9\%, depending on whether the PCA has been applied. The highest accuracy of 97\% has been obtained using stacked classifiers but at a cost of significantly longer training time. The accuracy achieved in this project should provide adequate support to SAXS analysts.

We believe that the tool is useful for both novice and experienced users. With some of the most common models in our training data, the SCAN tool can accelerate the model selection process already now. Furthermore, the training data can easily be expanded by including additional models from sasmodel and/or from labelled experimental data provided by users, making it easily adaptable to specific needs of individual analysts. At the same time the open source implementation makes it possible for anyone to contribute to the tool. The SCAN tool is created in such a way that it would be even possible to directly connect to an experimental setup to read the data file in real time. Thus, we conclude that the application of ML classifiers to the data generated by SAXS-measurements yields a route towards an accelerated handling of this complex class of data. 

\section*{Acknowledgment}
This initiative received financial support through three internal RISE initiatives, i.e., ``Machine Learning and Materials Design'', ''SODA - Software \& Data Intensive Applications'' and ``MLOps by RISE.'' This work benefited from the use of the SasView application, originally developed under NSF award DMR-0520547. SasView contains code developed with funding from the European Union’s Horizon 2020 research and innovation programme under the SINE2020 project, grant agreement No 654000.

\bibliographystyle{IEEEtran}
\bibliography{refs}

\end{document}